\documentclass[runningheads]{llncs}
\usepackage{graphicx}
\usepackage{amsmath,amssymb} 
\usepackage{multirow}
\usepackage{color}
\usepackage{ifthen}
\usepackage{esvect}
\usepackage{graphicx}
\usepackage{tabularx}
\usepackage{soul}
\usepackage{mathtools}
\usepackage{enumerate}
\usepackage{caption}
\usepackage{makecell}
\begin{document}
\pagestyle{headings}
\mainmatter

\definecolor{mygray}{gray}{.9}
\definecolor{gray}{rgb}{0.5,0.5,0.5} 
\definecolor{green}{rgb}{0, 0.4, 0} 
\definecolor{orange}{rgb}{1, 0.5, 0}	
\definecolor{mahogany}{rgb}{0.75, 0.25, 0.0}
\definecolor{purple}{rgb}{0.6, 0, 0.6}
\definecolor{darkgreen}{rgb}{0, 0.4, 0.4} 
\definecolor{frenchblue}{rgb}{0.0, 0.45, 0.73}
\definecolor{myblue}{RGB}{74, 120, 255}
\definecolor{mygreen}{RGB}{0, 180, 100}
\definecolor{red}{RGB}{255, 60, 60}

\newboolean{revising}
\setboolean{revising}{False}
\ifthenelse{\boolean{revising}}
{
	\newcommand{\ignore}[1]{}
    \newcommand{\Huang}[1]{\textcolor{black}{#1}}
	\newcommand{\Hsu}[1]{\textcolor{black}{#1}}
	\newcommand{\Chiu}[1]{\textcolor{black}{#1}}
	\newcommand{\Sun}[1]{\textcolor{black}{#1}}
    \newcommand{\New}[1]{\textcolor{red}{#1}}
} {
	\newcommand{\ignore}[1]{}
    \newcommand{\Huang}[1]{#1}
	\newcommand{\Hsu}[1]{#1}
	\newcommand{\Chiu}[1]{#1}
	\newcommand{\Sun}[1]{#1}
    \newcommand{\New}[1]{#1}
}

\title{Efficient Uncertainty Estimation for Semantic Segmentation in Videos} 

\titlerunning{Efficient Uncertainty Estimation for Semantic Segmentation in Videos}

\authorrunning{Po-Yu Huang, Wan-Ting Hsu, Chun-Yueh Chiu, Ting-Fan Wu, Min Sun}

\author{Po-Yu Huang\textsuperscript{1}, Wan-Ting Hsu\textsuperscript{1}, Chun-Yueh Chiu\textsuperscript{1}, Ting-Fan Wu\textsuperscript{2}, Min Sun\textsuperscript{1}}


\institute{\textsuperscript{1} National Tsing Hua University, \textsuperscript{2} Umbo Computer Vision\\
	\email{\{andy11330,cindyemail0720\}@gmail.com   chiupick86@gapp.nthu.edu.tw   tingfan.wu@umbocv.com   sunmin@ee.nthu.edu.tw}
}

\maketitle

\begin{abstract}
Uncertainty estimation in deep learning becomes more important recently. A deep learning model can't be applied in real applications if we don't know whether the model is certain about the decision or not. Some literature proposes the Bayesian neural network which can estimate the uncertainty by Monte Carlo Dropout (MC dropout). However, MC dropout needs to forward the model $N$ times which results in $N$ times slower. For real-time applications such as a self-driving car system, which needs to obtain the prediction and the uncertainty as fast as possible, so that MC dropout becomes impractical. In this work, we propose the region-based temporal aggregation (RTA) method which leverages the temporal information in videos to simulate the sampling procedure. \New{Our RTA method with Tiramisu backbone is \textbf{10x} faster than the MC dropout with Tiramisu backbone ($N=5$). Furthermore, the uncertainty estimation obtained by our RTA method is comparable to MC dropout's uncertainty estimation on pixel-level and frame-level metrics.}

\end{abstract}

\section{Introduction}


Nowadays, deep learning has become a powerful tool in various applications. The uncertainty estimation in deep learning has got more attention as well.
Some applications need not only the prediction of the model but also the confidence of this prediction. For instance, in the biomedical field, the confidence of cancer diagnosis is essential for doctors to make the decision. For self-driving car system to avoid accidents, the model should know what situation haven't seen before and then return to human control. There are some methods of uncertainty estimation \cite{kendall2015bayesian,gal2015bayesian,blundell2015weight} for deep learning have been proposed, but most of them need to sample several times, which is harmful to real-time applications. Slow inference of uncertainty estimation is an important issue before applying on real-time applications.

\begin{figure}
\begin{center}   
\includegraphics[width=1\linewidth]{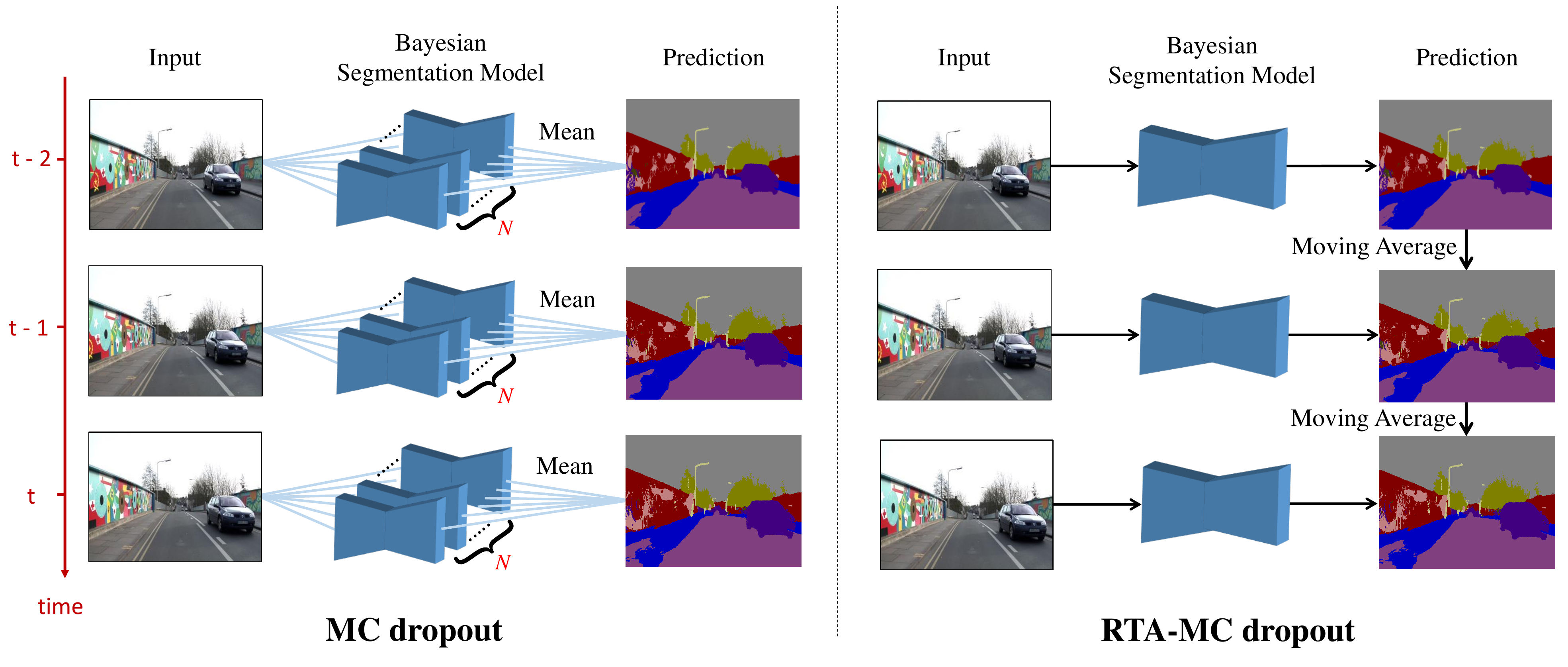}
\small \caption{\small Comparison of MC dropout and TA-MC/RTA-MC dropout. Left: MC dropout samples $N$ times for every frame, which cause N times slower. Right: TA-MC/RTA-MC dropout aggregates previous output into final prediction and uncertainty. For every frame, TA-MC/RTA-MC dropout only needs to calculate segmentation model and optical flow once.}
\label{fig.MC_vs_RATMC}
\end{center}
\end{figure}

In general, neural networks can only generate prediction instead of uncertainty. Lack of uncertainty estimation is a shortcoming of neural networks. Bayesian neural networks \cite{mackay1992practical,denker1991transforming} solve this problem by modeling the posterior of networks weights. But they often increase computation cost and the number of model parameters. Recently, Gal et al.\cite{gal2016dropout} propose dropout as an approximation technique without increasing parameters which is easy to implement called MC dropout. Though MC dropout is useful and powerful, the inference is very slow because it needs to perform $N$ (e.g., $N=50$) stochastic forward pass through the network and average the results to obtain the prediction and uncertainty. Therefore, our work proposed utilizing video's temporal information to speed up the inference and also maintain the performance.

For video segmentation, we can make good use of the temporal information based on the properties of video continuity.
We propose two main methods called \textit{\textbf{temporal aggregation (TA)}} and \textit{\textbf{region-based temporal aggregation (RTA)}}. For static objects in videos, calculating the average output of $N$ consecutive frames has the same effect as utilizing MC dropout with $N$ samples. Hence, we propose \textit{TA} method that approximates the sampling procedure of MC dropout by calculating the moving average of the outputs in consecutive frames (see Fig.~\ref{fig.MC_vs_RATMC}). To obtain the correct aggregation for moving objects in videos, we utilize optical flow to catch the flow of each pixel in the frame and aggregate each pixel's output depending on the flow. This \textit{TA} method can also be used to calculate any kinds of uncertainty estimation function, i.e, \textit{Entropy}, \textit{Bald}. \New{In this way, we can speed up MC dropout 10 times. The specific speed up rate is depend on backbone model. For larger backbone, our method can speed up even more.} Furthermore, we designed \textit{RTA} based on \textit{TA}. For some objects with large displacements in videos, the large shift of pixels might result in poor flow estimation and lead to wrong prediction and uncertainty estimation. Thus, \textit{RTA} can dynamically assign multiplying factor, which is used to decide the weight of incoming data, depending on the reconstruction error for every pixel. For pixels that have large reconstruction error, we shall assign higher multiplying factor so that they will rely more on themselves rather than the previous prediction. With the benefits of \textit{RTA}, we can get better prediction and uncertainty estimation.

In this paper, we mainly contribute three points:
\begin{itemize}
\item We propose \textit{\textbf{temporal aggregation (TA)}} method to solve the slow speed problem of MC dropout. We speed up more than 10 times comparing with MC dropout.
\item We propose \textit{\textbf{region-based temporal aggregation (RTA)}} method to further improve the performance of \textit{TA} by considering the flow accuracy. With our \textit{RTA} method, we get comparable accuracy in video segmentation on CamVid dataset with only less than 2\% drop on mean IoU metric.
\item We obtain nice uncertainty estimation which is evaluated in pixel-level and frame-level metric. Our uncertainty estimation even outperforms MC dropout on frame-level metrics.
\end{itemize}
 
\section{Related Work}

\Huang{First, We will introduce uncertainty estimation methods in Sec.~\ref{sec.RW_unct}. Next, some important segmentation models will be mentioned in Sec.~\ref{sec.RW_seg}. Finally, we introduce some works leverage the temporal information in the video.}

\subsection{Uncertainty Estimation}
\label{sec.RW_unct}
\Chiu{Uncertainty is an important issue for some current decision-making tasks, i.e., self-driving car, drone, robotics. We can just blindly assume that the prediction of the model is accurate but sometimes the truth is not. To really understand what a model doesn't know is a critical issue nowadays. It helps us to know how much we can trust the prediction of the model. However, the majority of the segmentation works cannot generate a probabilistic output with a measure of model uncertainty.}

\Huang{Bayesian neural networks \cite{mackay1992practical,denker1991transforming} is a well-known method that model uncertainty in neural networks. They turn deep learning model into a probabilistic model by learning the distribution over networks weights. Bayesian neural network's prediction is hard to obtain. Variational inference \cite{graves2011practical} is often used to approximate the posterior of the model. Blundell et al. \cite{blundell2015weight} model a Gaussian distribution over weights in the neural networks rather than having a single fixed value, however, each weight should contain mean and variance to represent a Gaussian distribution that doubles the number of parameters. Recently, Gal et al. \cite{gal2016dropout,gal2015bayesian,kendall2015bayesian} use dropout as an approximation of variational inference. When testing time, they keep dropping neurons, which can be interpreted as adding a Bernoulli distribution over the weights. This technique called MC dropout that has been successfully used in camera relocalisation \cite{kendall2016modelling} and segmentation \cite{kendall2015bayesian}. However, it still needs to sample model many times to estimate uncertainty. In this work, we propose leveraging video temporal information to speed up the MC dropout sampling process.}

\subsection{Semantic Segmentation}
\label{sec.RW_seg}
Semantic image segmentation that uses convolutional neural networks has achieved several breakthroughs in recent years. 
It is a pixel-wise labeling task that classifies every pixel into defined class. Long et al. \cite{long2015fully}, popularize CNN architectures for dense predictions without any fully connected layers. This method allowed segmentation maps to be generated for an image of any size and was also much faster compared to the patch classification approach. Ronneberger et al. \cite{ronneberger2015u} propose U-net, which is an encoder-decoder architecture that focuses on improving more accurate boundaries. Howard et al.\cite{howard2017mobilenets} combined the ideas of MobileNets Depthwise Separable Convolutions with UNet to build a high speed, low parameter Semantic Segmentation model.  PSP-Net \cite{zhao2017pyramid} uses ResNet as the backbone and utilizes global information from pyramid layers to provide more accurate semantics. DeepLab \cite{chen2016deeplab} replaced fully connected CRF(conditional random field) to the last layer of CNN for improving the performance. \New{In this work, we select Bayesian SegNet \cite{badrinarayanan2017segnet} and Tiramusi \cite{jegou2017one} to demonstrate our idea. Both methods are encoder-decoder architecture. Tiramisu is the state-of-the-art of CamVid dataset. }

\subsection{Leverage Temporal Information}
\Chiu{Previously, some works make use of superpixels \cite{chang2013video,grundmann2010efficient} , patches \cite{fan2015jumpcut,ramakanth2014seamseg}, object proposal \cite{perazzi2015fully}, optical flow \cite{jang2017online,perazzi2017learning} as temporal information to reduce the computational complexity. Furthermore, video segmentation has gained significant improvement based on temporal information. Among all these temporal information, the most recent works heavily rely on optical flow. Srivastava et al.\cite{srivastava2014dropout} use the image in one stream, and optical flow in the other stream to recognize actions in the video. Simonyan et al.\cite{simonyan2014two} simultaneously predict pixel-wise object segmentation and optical flow in videos. Cheng et al.\cite{cheng2017segflow} emphasize temporal information at the frame level instead of the final box level to improve detection accuracy. To enhance the reference feature map, they utilize optical flow network the work of Zhu et al.\cite{zhu2017flow} to estimate the motions between nearby frames and the reference frame. They then aggregate feature maps warping from nearby frames to the reference frame according to the flow motion. Briefly speaking, all these works utilize optical flow appropriately in video tasks. To the best of our knowledge, we are the first work that uses optical flow as temporal information to speed up uncertainty estimation. }

\section{Method}
\label{sec:blind}

\Hsu{We first give a brief introduction of Bayesian neural networks with Monte Carlo dropout (MC) in Sec.~\ref{sec.pre}. Next, we introduce our temporal aggregation Monte Carlo dropout (TA-MC) in Sec.~\ref{sec.TAMC}. Finally, we propose a region-based temporal aggregation Monte Carlo dropout (RTA-MC) which can further improve both the accuracy and uncertainty estimation in Sec.~\ref{sec.RTAMC}.}

\subsection{Preliminary: Bayesian Neural Network with MC dropout}
\label{sec.pre}


\Huang{Bayesian neural networks are probabilistic models that do not learn a set of deterministic parameters but a distribution over those parameters. It aims to learn the posterior distribution of the neural network's weights $W$ given training data $X$ and $Y$.}

\Huang{The posterior distribution, which is denoted as $p\left ( W|X,Y \right )$, usually cannot be evaluated analytically. Variational inference is often used to approximate the posterior distribution. Given an approximating distribution over the network's weights, $q\left( W\right)$, we minimize the Kullback-Leibler (KL) divergence between $p\left ( W|X,Y \right )$ and $q\left( W\right)$.}
\Huang{\begin{align}
KL\left ( q\left ( W \right )\parallel p\left ( W|X,Y \right ) \right )
\end{align}
Dropout variational inference is a useful technique for approximating posterior distribution. \New{Dropout can be viewed as using the Bernoulli distribution as the approximation distribution $q\left ( W \right )$.} At testing time, the prediction can be approximated by sampling model $N$ times which is referred as Monte Carlo dropout (MC).}

\Huang{\begin{align}
p\left (y^{*}| x^{*}, X, Y \right ) \approx \frac{1}{N}\sum_{n=1}^{N} p\left (y^{*}| x^{*}, \hat{\omega_{n}}  \right )
\end{align}
The uncertainty of the classification can be obtained by several functions: \\
\begin{enumerate}[(a)]
\item Entropy \cite{shannon2001mathematical}: 
\begin{align}H\left [ y|x, X, Y \right ] = -\sum_{c}^{ }p\left ( y=c|x, X, Y \right )\log p\left ( y=c|x, X, Y \right )\end{align}
\item BALD \cite{houlsby2011bayesian}:
\begin{align}I\left [ y, \omega |x, X, Y \right ] = H\left [ y|x, X, Y \right ]-E_{p\left ( \omega |X,Y \right )}\left [ H\left [ y|x,\omega \right ] \right ]\end{align}
\item Variation ratio \cite{freeman1965elementary}:
\begin{align}\text{variation-ratio}\left [ x \right ] = 1-\max_{y}p\left ( y|x,X,Y \right )\end{align}
\item Mean standard deviation (Mean STD) \cite{kampffmeyer2016semantic,kendall2015bayesian}:
\begin{align}\sigma_{c}=\sqrt{E_{q\left ( \omega \right )}\left [ p\left ( y=c|x,\omega \right )^{2} \right ] - E_{q\left ( \omega \right )}\left [ p\left ( y=c|x,\omega \right ) \right ]^{2}}\end{align}
\begin{align}\sigma \left ( x \right )=\frac{1}{c}\sum_{c}^{ }\sigma_{c}\end{align}
\end{enumerate}}

\Hsu{Bayesian neural networks with MC dropout can obtain better performance and uncertainty estimation. However, it requires to sample $N$ times (e.g., $N = 50 $) for predicting each image, which is $N$ times slower than the original network. For real-time applications such as self-driving cars, which needs to obtain the prediction and uncertainty estimation as fast as possible, so that MC dropout becomes impractical. In this work, we propose temporal aggregation MC dropout to speed up the MC dropout process.}

\begin{figure}[t]
\begin{center}   
\includegraphics[width=1 \textwidth]
{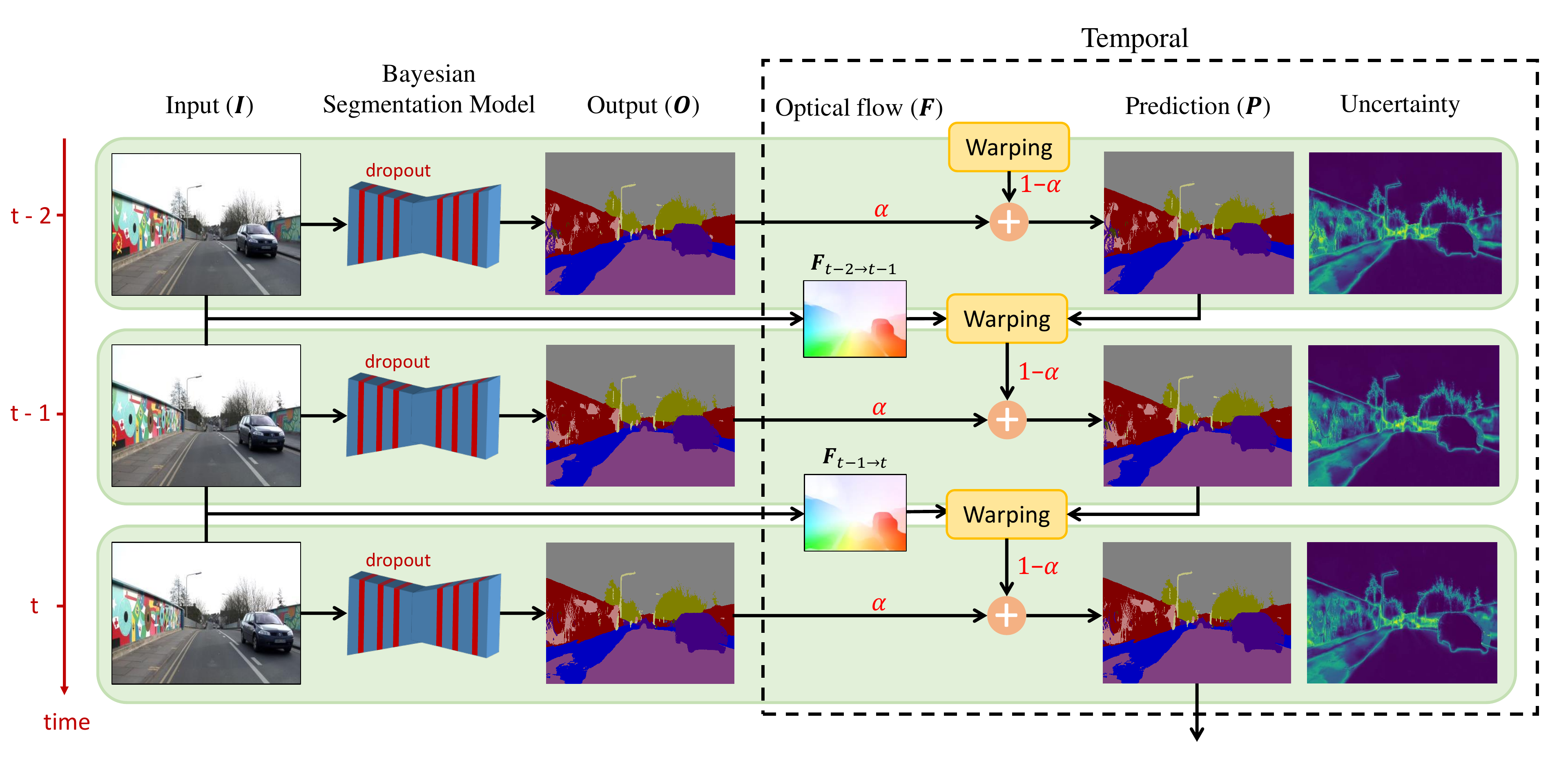}
\small \caption{\small The process of our temporal aggregation MC dropout (TA-MC) for video segmentation. Every time step Bayesian segmentation model sample one output and calculate optical flow. To average the incoming output, we warp the previous prediction depend on optical flow. Final prediction is weighted sum with multiplying factor $\boldsymbol{\alpha}$.}
\label{fig.TAMC}
\end{center}
\end{figure}

\subsection{Temporal Aggregation MC Dropout (TA-MC)}
\label{sec.TAMC}



\Hsu{Our temporal aggregation MC dropout (TA-MC) method utilizes the temporal property in videos. Since a video contains consecutive frames, same objects may appear in many different frames and thus will be forwarded by the Bayesian model repeatedly. If a video contains static frames (i.e., a static scene observed by a static camera), the average output of $N$ consecutive frames is the same as MC dropout with $N$ samples.}
\Hsu{For video segmentation, though the frames are not static (i.e., the objects in the scene and the camera are both moving), the consecutive frames are still similar. The objects are often shifted slightly in the next frame. Hence, by warping each pixel to the new position in the next frame, we can aggregate the outputs of the pixels in consecutive frames correctly.}

\noindent\textbf{Notations.} 
Given a video $\boldsymbol{V} = \left \{ I_{1}, I_{2}, ..., I_t, ..., I_{T} \right \}$ where $I_t$ is the $t^{th}$ frame and $T$ is the length of the video, the outputs of the Bayesian neural network are denoted as $\boldsymbol{O} = \left \{ O_{1}, O_{2}, ..., O_t, ..., O_{T} \right \}$. Note that $\boldsymbol{O}$ are the outputs without MC sampling, \New{which means each frame is forwarded by Bayesian model (with dropout) for only one time.} To get the aggregated predictions, we calculate the optical flow between consecutive frames $\boldsymbol{F} = \left \{ F_{1\rightarrow 2}, F_{2\rightarrow 3}, ..., F_{T-1\rightarrow T} \right \}$ where $F_{t\rightarrow t+1}$ indicates the optical flow from frame $I_t$ to frame $I_{t+1}$.

\noindent\textbf{Aggregated Prediction.}
\Hsu{The prediction $P_t$ for each frame $I_t$ is obtained by calculating the weighted moving average of the outputs $\boldsymbol{O}_{1:t}=\{O_1, O_2, ..., O_{t}\}$:
\begin{equation}
P_{t}=
\begin{cases}
O_{t} & \text{if } t =1,\\
O_{t} \times \alpha + W\left ( P_{t-1}, F_{t-1\rightarrow t} \right ) \times (1-\alpha) & \text{otherwise},\\
\end{cases}
\label{eq.mean}
\end{equation}
where $W\left (\cdot \right ) $ is a pixel-wise warping function that moves the input values (e.g., $P_{t-1}$) to their new positions depending on the given optical flow (e.g., $F_{t-1}$). The output of $W(\cdot)$ has the same dimension as the input. $\alpha$ is a multiplying factor which decides the weights of the incoming data and previous data. The whole system of our TA-MC dropout for video segmentation is shown in Fig.~\ref{fig.TAMC}.}

\noindent\textbf{Uncertainty Estimation.}
\Huang{Our temporal aggregation method can be used to calculate any kinds of uncertainty estimation mentioned in Sec.~\ref{sec.pre}. For \textit{entropy} and \textit{variation ratio}, the uncertainty can be simply derived from aggregated prediction $P_{t}$, while \textit{BALD} and \textit{Mean STD} encounter other expectation that needed to be aggregated. \textit{BALD} needs to aggregate $E_{p\left ( \omega |X,Y \right )}\left [ H\left [ y|x,\omega \right ] \right ]$ and \textit{Mean STD} needs to aggregate $E_{q\left ( \omega \right )}\left [ p\left ( y=c|x,\omega \right ) \right ]^{2}$.} 

\begin{figure}[t]
\begin{center}   \includegraphics[width=1\linewidth]{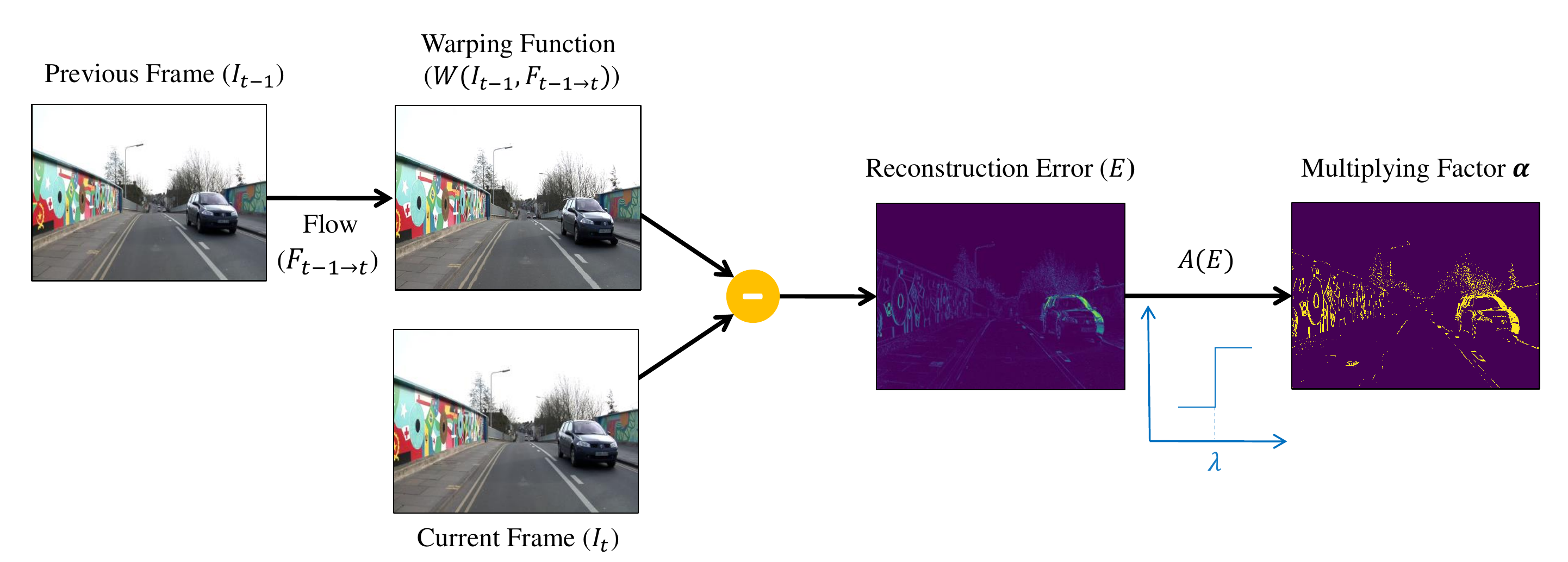}
\end{center}
\small \caption{\small Region-based temporal aggregation (RTA). We design a step function to acquire dynamic multiplying factor $\boldsymbol{\alpha}$ for improving the TA method. For regions that have wrong optical flow estimation (i.e., the reconstruction error is greater than a threshold $\lambda$), we use a larger multiplying factor to let the pixels rely more on itself rather than the previous predictions. See more detailed in Sec.~\ref{sec.RTAMC}. }
\label{fig.RTAMC}
\end{figure}

\subsection{Region-Based Temporal Aggregation MC Dropout (RTA-MC)}
\label{sec.RTAMC}

\Hsu{Our TA-MC dropout works for most of the case; however, when the optical flow estimation is wrong, it will cause the uncertainty estimation inaccurate. For some regions that contain fast moving objects or occlusion, the optical flow may not be accurate. Bad flow estimation results in calculating moving average on the wrong patch and thus getting wrong prediction and uncertainty for those pixels. To solve this problem, we propose the region-based temporal aggregation (RTA) that can dynamically assign different multiplying factor ($\alpha$ in Eq.~\ref{eq.mean}) for every pixel depending on its reconstruction error.}

\Hsu{ The reconstruction error $E$ of warping $I_{t-1}$ to $I_t$ is derived from the pixel-wise difference of the warped frame $W\left ( I_{t-1}, F_{t-1\rightarrow t} \right )$ and $I_t$.
\begin{equation}
E= |I_{t} - W\left (I_{t-1}, F_{t-1\rightarrow t}\right )|.
\end{equation}
where $E$ is a matrix contains pixel-wise reconstruction error and $E_{ij} \in [0,255]$. For a pixel that has large reconstruction error, we will give it a higher multiplying factor $\alpha_{err}$ since the optical flow may be inaccurate and thus the prediction should rely more on itself rather than the previous predictions. We design a decision function $A \left ( E \right )$ that decides $\alpha$ for every pixel depending on the reconstruction error (see Fig.~\ref{fig.RTAMC}).
\begin{equation}
\begin{split}
&\alpha_{ij}=A\left ( E_{ij} \right )=
\begin{cases}
\alpha_{acc} & \text{if } E_{ij} \leq \lambda,\\
\alpha_{err}  &  \text{otherwise},\\
\end{cases}
\\
&\boldsymbol{\alpha}=\{\alpha_{ij}\}_{ij}
\end{split}
\label{eq.spatial_alpha}
\end{equation}
where $\alpha_{ij}$ is the multiplying factor for the pixel in position $(i,j)$. $A(\cdot)$ is a step function with a threshold $\lambda$ which is a hyper-parameter deciding whether the optical flow has bad estimation or not. $\alpha_{acc}$ and $\alpha_{error}$ are also hyper-parameters which indicate the multiplying factor for good flow estimation and bad flow estimation, respectively. $\alpha_{err}$ should be higher than $\alpha_{acc}$ (e.g., $\alpha_{acc}=0.2$ and $\alpha_{err}=0.7$). Then, we simply replace $\alpha$ in Eq.~\ref{eq.mean} with $\boldsymbol{\alpha}$ in Eq.~\ref{eq.spatial_alpha} to obtain our region-based temporal aggregation MC dropout (RTA-MC). By applying our RTA method, the mismatch aggregation will be attenuated and can further improve the prediction and the uncertainty estimation.}


\section{Experiment}

\Hsu{In this section, we describe the dataset that used for the video segmentation in Sec.~\ref{sec.dataset} and our implementation details in Sec.~\ref{sec.impl}. Next, we compare our TA-MC dropout and RTA-MC dropout with MC dropout in several different aspects. First, in Sec.~\ref{sec.acc}, we compare the performance of video segmentation and the inference time. Second, in Sec.~\ref{sec.unct},  we show that our methods can obtain comparable uncertainty estimation.}

\subsection{Dataset}
\label{sec.dataset}
\Hsu{CamVid \cite{brostow2009semantic} is a road scene segmentation dataset which contains four 30Hz videos. The frames are labeled every 1 second, and each pixel is labeled into 11 classes such as sky, building, road, car, etc. There are total 701 labeled frames split into 367 training frames, 101 validation frames and 233 test frames.  All frames are resized to 360x480 pixels in our experiments.}

\subsection{Implementation Details}
\label{sec.impl}

\New{We apply the Bayesian SegNet model \cite{kendall2015bayesian} and Tiramisu model \cite{jegou2017one} to demonstrate our TA and RTA method. We train both models by the same setting in the original paper.} We set $\alpha$ in Eq.~\ref{eq.mean} for TA-MC dropout to 0.2. For RTA-MC dropout, we use threshold $\lambda$ as 10 to determine whether the flow estimation is wrong. Note that the reconstruction error of flow is in the range of $[0,255]$ and the average error is about 2. Hence, we choose the threshold $\lambda$ slightly higher than the average error. $\alpha_{acc}$ and $\alpha_{err}$ in Eq.~\ref{eq.spatial_alpha} are set to 0.2 and 0.7, respectively. We use FlowNet 2.0 \cite{ilg2017flownet} for our optical flow estimation. We implement all methods in Pytorch \cite{paszke2017automatic} framework and experiment on GTX 1080 for time measurement.

\begin{table}[t]
\centering
\setlength{\extrarowheight}{0.05mm}
\addtolength{\tabcolsep}{0.5mm}
\fontsize{6pt}{10pt}\selectfont
\begin{tabular}{ccccccccccccccccc}
Method & \rotatebox{90}{Building} & \rotatebox{90}{Tree} & \rotatebox{90}{Sky} & \rotatebox{90}{Car}    & \rotatebox{90}{Sign-Symbol} & \rotatebox{90}{Road}  & \rotatebox{90}{Pedestrian} & \rotatebox{90}{Fence} & \rotatebox{90}{Column-Pole} & \rotatebox{90}{Side-walk} & \rotatebox{90}{Bicyclist} & \rotatebox{90}{Class avg.} & \rotatebox{90}{Global avg.} & \rotatebox{90}{Mean IoU} & \rotatebox{90}{Inference Time (s)}& \rotatebox{90}{Speed Up Ratio}\\ 
\hline\hline

\multicolumn{1}{c|}{\begin{tabular}[c]{@{}c@{}}SegNet \\ MC (N=50) \end{tabular}} & \multicolumn{1}{c|}{88.9}   & \multicolumn{1}{c|}{88.7}     & \multicolumn{1}{c|}{95.0}     & \multicolumn{1}{c|}{89.0}   & \multicolumn{1}{c|}{49.4}     & \multicolumn{1}{c|}{95.8}     & \multicolumn{1}{c|}{73.5}     & \multicolumn{1}{c|}{51.4}     & \multicolumn{1}{c|}{43.3}     & \multicolumn{1}{c|}{93.2}     & \multicolumn{1}{c|}{57.0}     &     75.0       &    90.6         &    \multicolumn{1}{c|}{63.7} &2.00&1      \\ \hline
\multicolumn{1}{c|}{\begin{tabular}[c]{@{}c@{}}SegNet \\ TA-MC\end{tabular}}      & \multicolumn{1}{c|}{88.6}   & \multicolumn{1}{c|}{89.7}     & \multicolumn{1}{c|}{94.5}     & \multicolumn{1}{c|}{88.2}   & \multicolumn{1}{c|}{48.1}     & \multicolumn{1}{c|}{95.3}     & \multicolumn{1}{c|}{70.7}     & \multicolumn{1}{c|}{44.7}     & \multicolumn{1}{c|}{36.4}     & \multicolumn{1}{c|}{94.1}     & \multicolumn{1}{c|}{52.7}     &        73.0    &    90.3         &     \multicolumn{1}{c|}{62.1}  &0.18&10.97 \\ \hline
\multicolumn{1}{c|}{\begin{tabular}[c]{@{}c@{}}SegNet \\ RTA-MC\end{tabular}}     & \multicolumn{1}{c|}{88.4}   & \multicolumn{1}{c|}{89.3}     & \multicolumn{1}{c|}{94.9}     & \multicolumn{1}{c|}{88.9}   & \multicolumn{1}{c|}{48.7}     & \multicolumn{1}{c|}{95.4}     & \multicolumn{1}{c|}{73.0}     & \multicolumn{1}{c|}{45.6}     & \multicolumn{1}{c|}{41.4}     & \multicolumn{1}{c|}{94.0}     & \multicolumn{1}{c|}{51.6}     &     73.7       &     90.4        &    \multicolumn{1}{c|}{62.5}    &0.18&10.97   \\ \hline\hline
\multicolumn{1}{c|}{\begin{tabular}[c]{@{}c@{}}Tiramisu \\ MC (N=50) \end{tabular}} & \multicolumn{1}{c|}{89.7}   & \multicolumn{1}{c|}{87.2}     & \multicolumn{1}{c|}{95.6}     & \multicolumn{1}{c|}{84.9}   & \multicolumn{1}{c|}{58.4}     & \multicolumn{1}{c|}{95.1}     & \multicolumn{1}{c|}{82.5}     & \multicolumn{1}{c|}{54.1}     & \multicolumn{1}{c|}{49.6}     & \multicolumn{1}{c|}{84.6}     & \multicolumn{1}{c|}{52.3}     &     75.8       &    89.8         &    \multicolumn{1}{c|}{64.0}    &11.72&1  \\ \hline
\multicolumn{1}{c|}{\begin{tabular}[c]{@{}c@{}}Tiramisu \\ MC (N=5) \end{tabular}} & \multicolumn{1}{c|}{88.7}   & \multicolumn{1}{c|}{86.6}     & \multicolumn{1}{c|}{95.4}     & \multicolumn{1}{c|}{83.7}   & \multicolumn{1}{c|}{58.4}     & \multicolumn{1}{c|}{94.6}     & \multicolumn{1}{c|}{80.6}     & \multicolumn{1}{c|}{52.0}     & \multicolumn{1}{c|}{49.2}     & \multicolumn{1}{c|}{84.0}     & \multicolumn{1}{c|}{55.0}     &     75.3       &    89.2         &    \multicolumn{1}{c|}{62.4}     &1.17&10.00 \\ \hline
\multicolumn{1}{c|}{\begin{tabular}[c]{@{}c@{}}Tiramisu \\ TA-MC\end{tabular}}      & \multicolumn{1}{c|}{90.3}   & \multicolumn{1}{c|}{87.4}     & \multicolumn{1}{c|}{94.8}     & \multicolumn{1}{c|}{84.2}   & \multicolumn{1}{c|}{55.8}     & \multicolumn{1}{c|}{94.5}     & \multicolumn{1}{c|}{79.2}     & \multicolumn{1}{c|}{51.3}     & \multicolumn{1}{c|}{40.6}     & \multicolumn{1}{c|}{85.6}     & \multicolumn{1}{c|}{46.7}     &        73.8    &    89.6         &     \multicolumn{1}{c|}{62.3} &0.37&31.58  \\ \hline
\multicolumn{1}{c|}{\begin{tabular}[c]{@{}c@{}}Tiramisu \\ RTA-MC\end{tabular}}     & \multicolumn{1}{c|}{90.1}   & \multicolumn{1}{c|}{87.1}     & \multicolumn{1}{c|}{94.9}     & \multicolumn{1}{c|}{84.1}   & \multicolumn{1}{c|}{56.7}     & \multicolumn{1}{c|}{94.7}     & \multicolumn{1}{c|}{79.2}     & \multicolumn{1}{c|}{48.1}     & \multicolumn{1}{c|}{42.2}     & \multicolumn{1}{c|}{85.4}     & \multicolumn{1}{c|}{49.8}     &     73.9       &     89.5        &    \multicolumn{1}{c|}{62.4}    &0.37&31.58   \\ \hline
\end{tabular}
\small \caption{\small \New{Performance test on CamVid dataset. Upper three rows are comparisons of SegNet backbone; Lower four rows are comparisons of Tiramisu backbone. Both comparisons show that our methods can speed up more than 10x with only 1-2 percentage drop. For fairly comparison, we reduce the Tiramisu MC sample time to N=5 to get the same accuracy as our methods. In this situation, our methods are still 10x faster.}}
\label{table.acc}
\end{table}

\subsection{Results of Video Segmentation}
\label{sec.acc}
\New{We compare MC dropout with our TA-MC dropout and RTA-MC dropout on CamVid dataset by two models.} We first show the performance of the video segmentation on several different metrics: (1) pixel-wise classification accuracy on every class, (2) class average accuracy (class avg.), (3) overall pixel-wise classification accuracy (global avg.), (4) mean intersection over union (mean IoU) and (5) inference time. The results are shown in Table~\ref{table.acc}. For SegNet, Our TA-MC dropout can reach comparable accuracy, and the RTA-MC dropout further improves the performance with only 1.2\% drop on mean IoU metric. \New{For Tiramisu, our methods also can reach comparable accuracy.}

\New{In the case of comparable accuracy, our method can further speed up the inference time. Since our methods only need to forward one time for each frame; while MC dropout needs to sample $N$ times. For SegNet, we can obtain almost 11 times speed up. Note that our TA-MC dropout and RTA-MC dropout can perform the same speed as the only difference between them is the multiplying factor ($\alpha$ in Eq.~\ref{eq.mean}) which doesn't affect the speed. The inference time of the RTA-MC dropout mainly contains the inference time of the Bayesian SegNet model and the FlowNet 2.0 model which are 0.04 seconds and 0.13 seconds, respectively. FlowNet 2.0 model takes 70\% of the whole inference time. If we use the bigger segmentation model, we can get a better improvement in the speed. Therefore, we use Tiramisu model which is the state-of-the-art model in CamVid but 6x slower than SegNet to show better speed up ratio. For Tiramisu, Our method can achieve 31x faster than MC dropout sample 50 times. To fairly compare inference time, we reduce the MC dropout sample time to 5 times. The accuracy becomes the same as our methods. In this case, our methods are still 10x faster than MC dropout. This table shows that in the same accuracy level, our methods can speed up inference time 10x.}




\begin{figure}[t]
\begin{center}   
\includegraphics[width=0.9\linewidth]{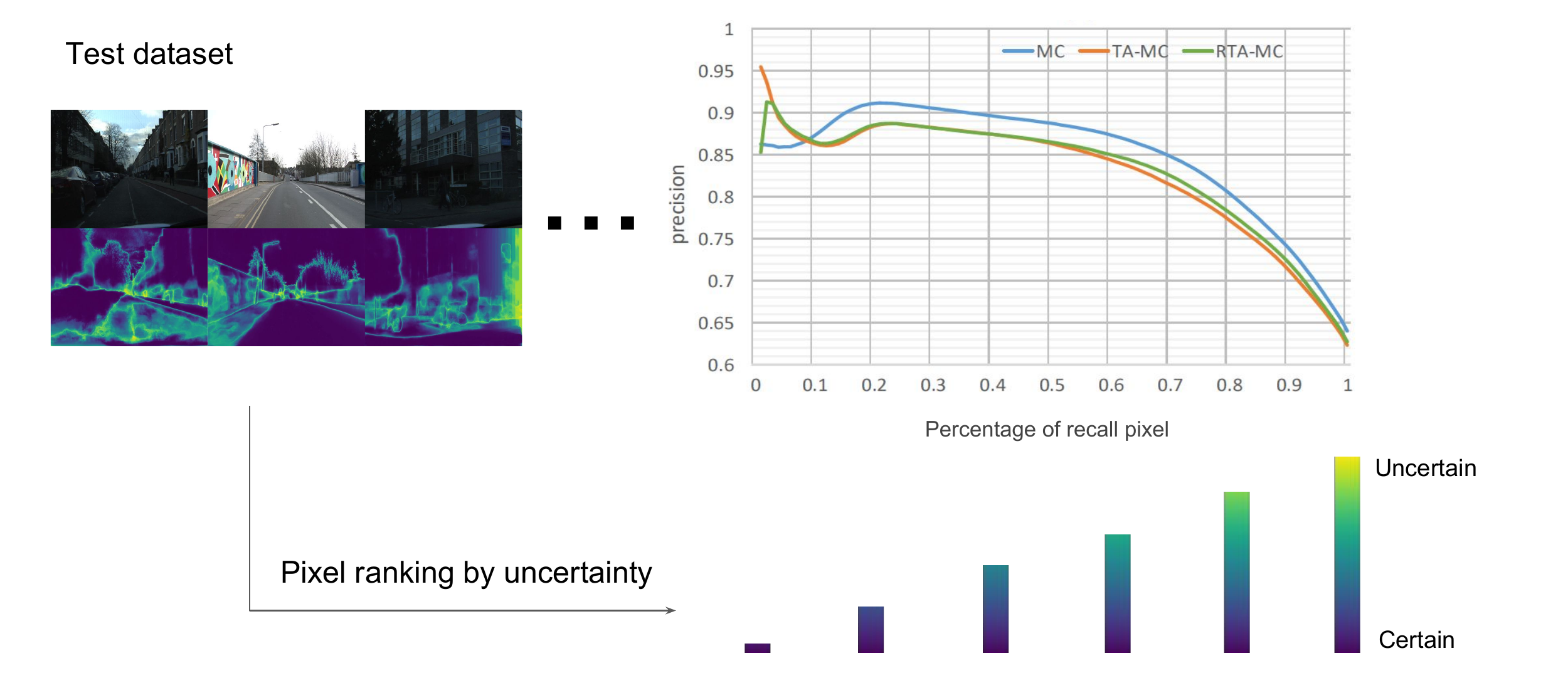}
\small \caption{\small \New{Explanation of the Pixel-level metric Precision-Recall curve. First, calculate the uncertainty map of all test data. Then rank all pixels by uncertainty value. The horExplanizontal axis is the percentage of recall pixel which means we keep how many percentages of most certain pixels to calculate precision. The vertical axis is the mIoU.}}
\label{fig.pr_explan}
\end{center}
\end{figure}





\subsection{Results of Uncertainty Estimation}

\begin{figure}[t]
\centering
\setlength\belowcaptionskip{-20pt}
\begin{minipage}{.5\textwidth}
  \centering
  \includegraphics[width=\linewidth]{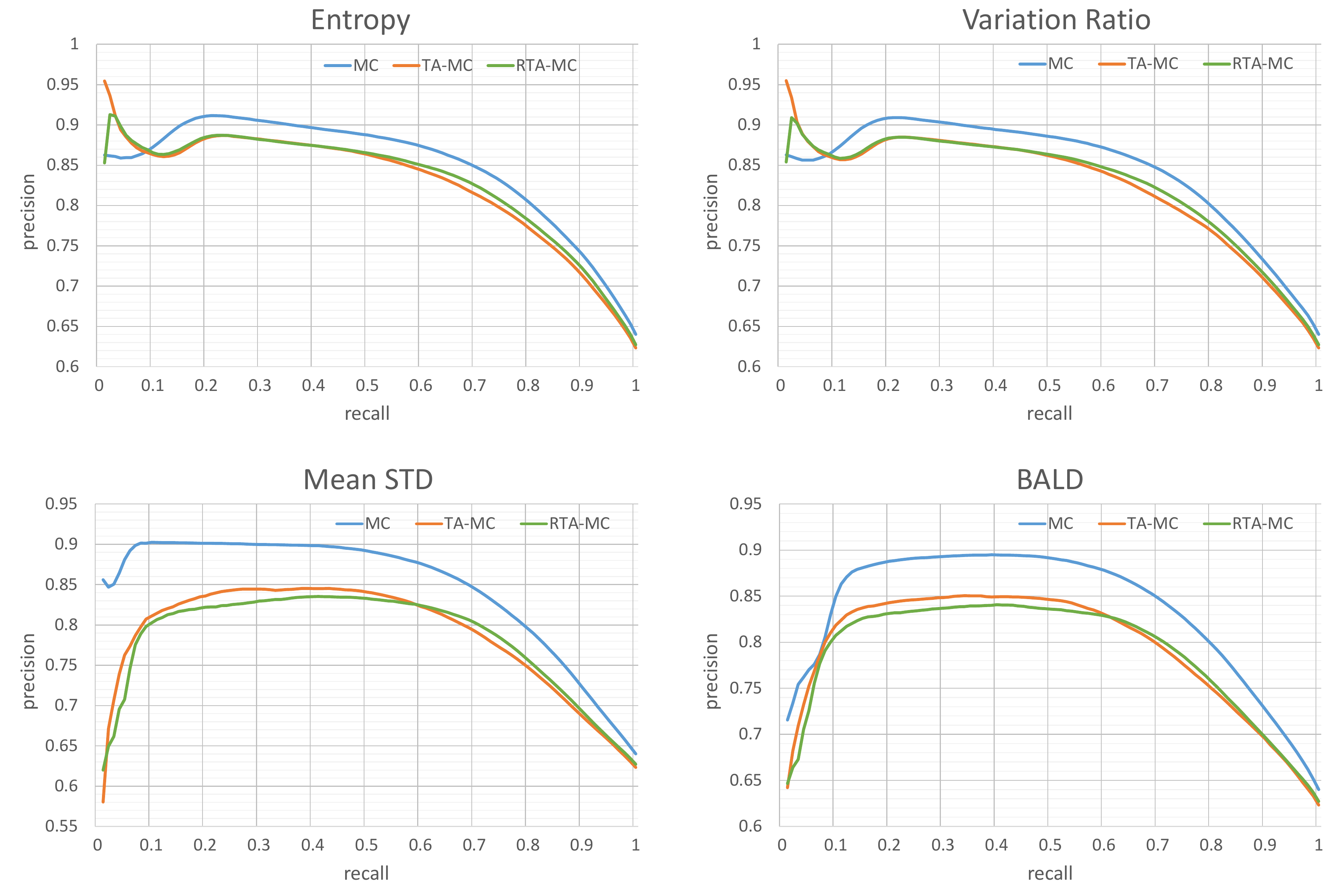}
\end{minipage}%
\begin{minipage}{.5\textwidth}
  \centering
  \includegraphics[width=\linewidth]{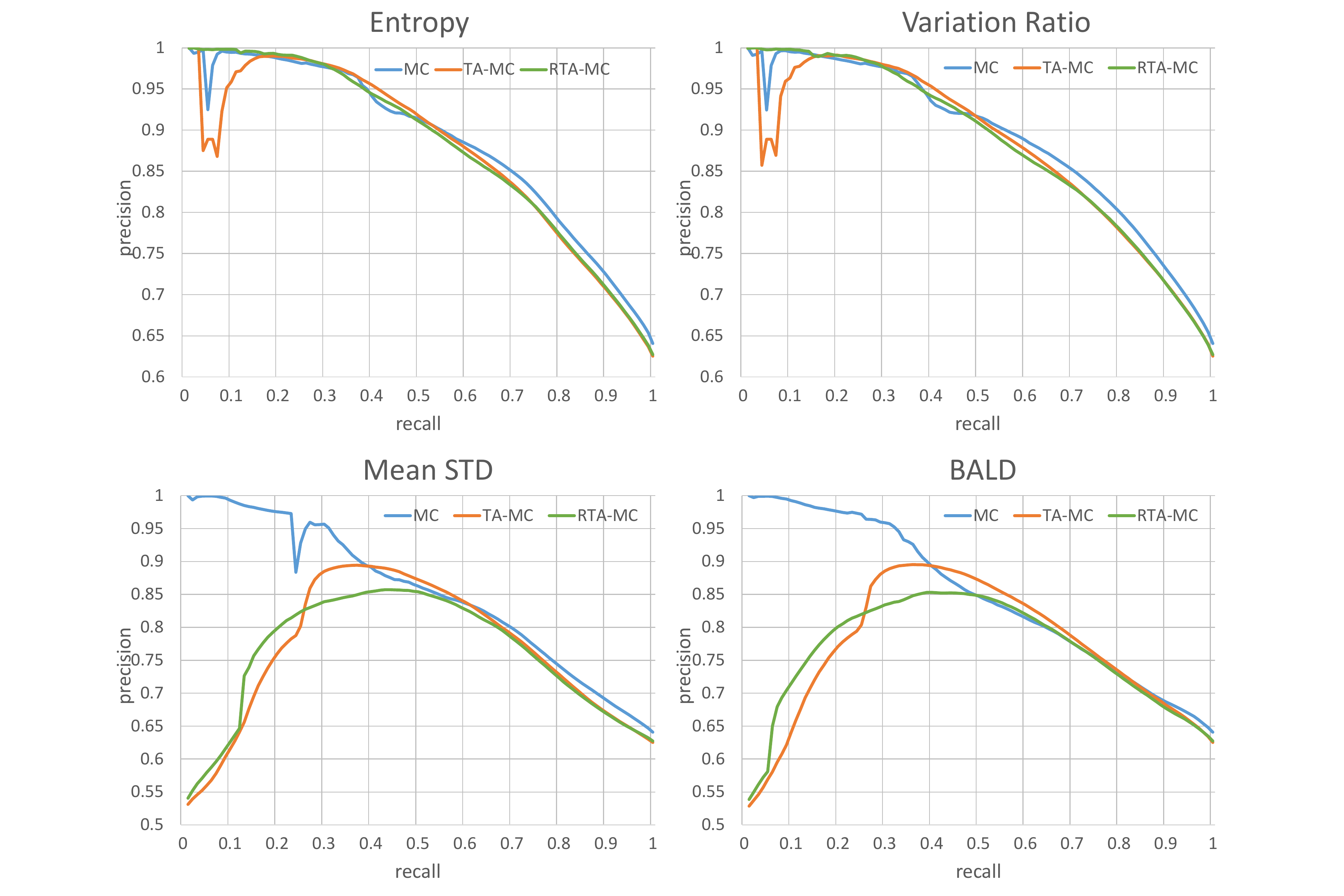}
\end{minipage}
\small \caption{\small \New{Pixel-level precision-recall curves. Left four figures are the results of SegNet backbone. Right four figures are the results of Tiramisu backbone. We use mean IoU as the precision metric. We show the comparison of MC dropout, TA-MC dropout and RTA-MC dropout on four different uncertainty estimation methods. Our methods achieve comparable results especially when using \textit{Entropy} and \textit{Variation Ratio} as the uncertainty estimation functions.}}
\label{fig.pr_curve}
\end{figure}

We evaluate the uncertainty estimation in pixel-level and frame-level metrics.\\
\noindent\textbf{Pixel-level Metric.}  The pixel-level evaluation is inspired by Precision-Recall Curve metric in \cite{kendall2017uncertainties}. This curve shows the accuracy of the remained pixels as removing pixels with uncertainty larger than different percentile thresholds. Detail explanation is in Fig.~\ref{fig.pr_explan}. A reliable uncertainty estimation should let the PR-Curve monotonically decrease. We compare MC dropout and our TA and RTA method with different uncertainty function in Fig.~\ref{fig.pr_curve}. Left four figures are the results of SegNet backbone. Right four figures are the results of Tiramisu backbone. All results show that as the recall percentage drop from 1 to 0.5, the mean IoU of all methods monotonically increase which means the uncertain pixels is correlated to misclassified pixels. Although MC dropout has the highest accuracy almost at all percentage, our TA-based methods are still comparable to MC dropout. TA-MC and RTA-MC have similar results in PR-Curve, but at the frame-level metric, RTA-MC will outperform TA-MC.\\

\begin{figure}[t]
\begin{center}   
\includegraphics[width=0.9\linewidth]{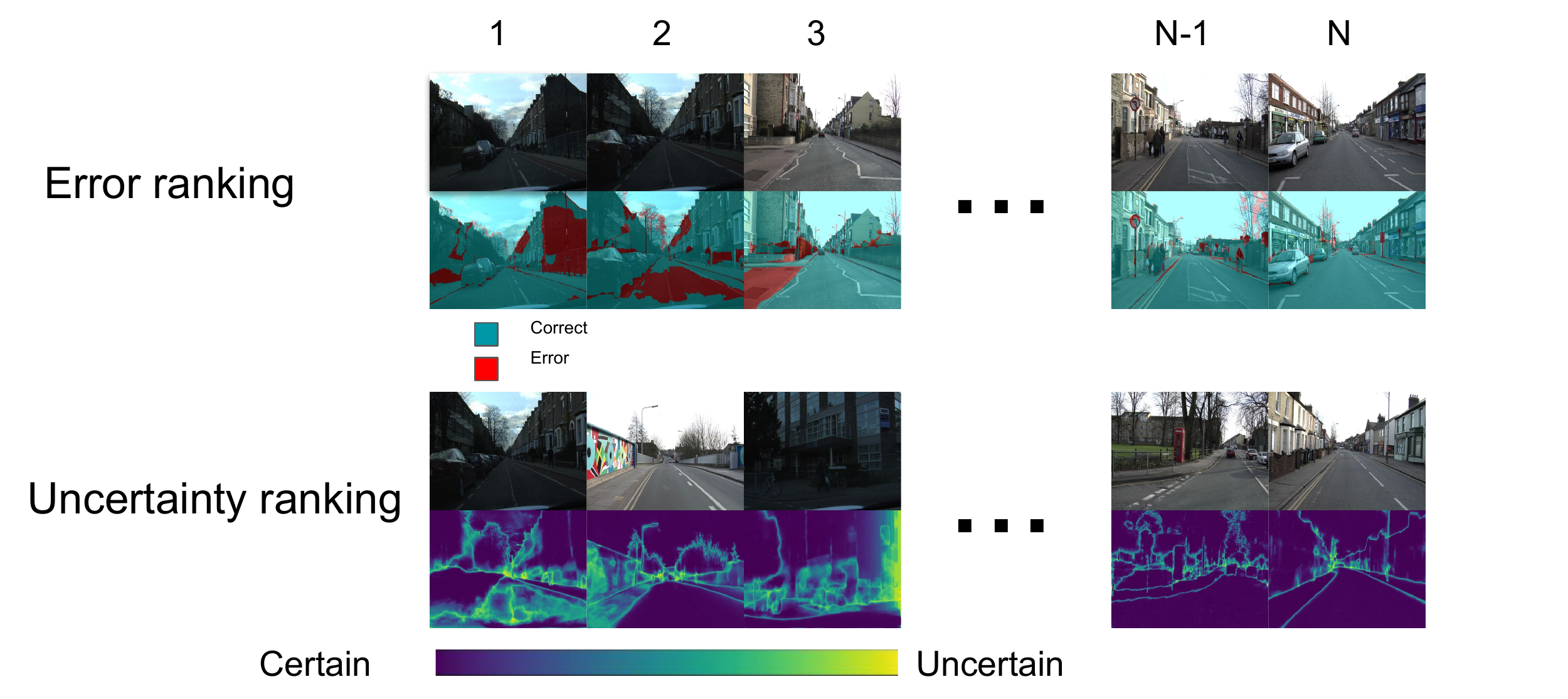}
\small \caption{\small \New{Explanation of frame-level metric. First, calculate the error rate of each test data frame by looking at the ground truth to get the error ranking sequence. Second, calculate the frame uncertainty of all test data to get the uncertainty ranking sequence. Then measure the similarity between two sequences by Kendall tau and Ranking IOU.}}

\label{fig.frame_level_explain}
\end{center}
\end{figure}

\begin{figure}
\begin{center}   
\includegraphics[width=0.9 \textwidth, height =0.7 \textheight]{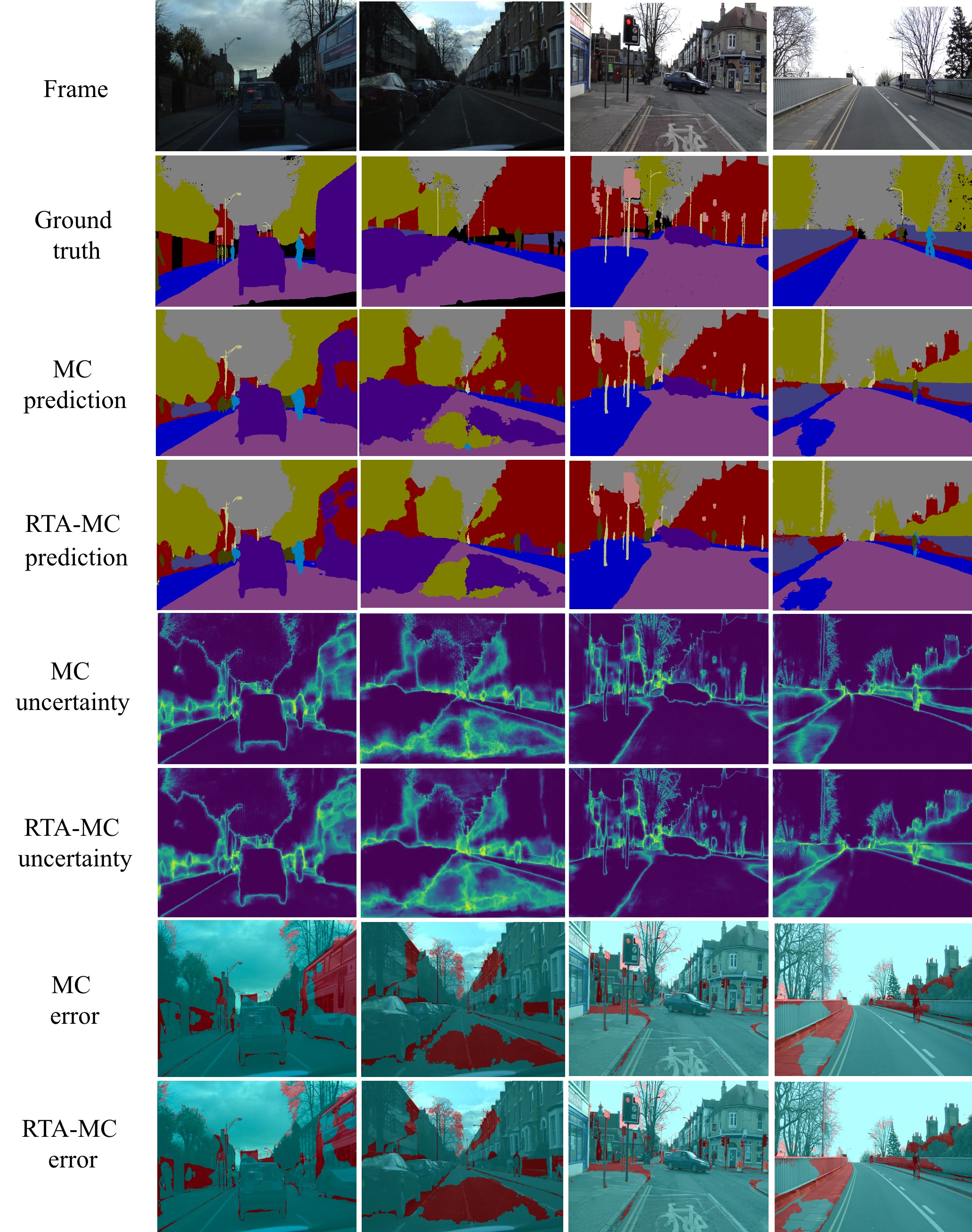}
\small \caption{\small Results comparison on CamVid dataset(MC dropout v.s RTA-MC dropout). The top row is the input image, with the ground truth shown in the second row. The third row and fourth row show the segmentation prediction of MC dropout and RTA-MC respectively. Its corresponding uncertainty map is also shown in the fifth and sixth row where the more brighter space represents higher uncertainty. We even show the error in the last two rows where the red space represents the wrong prediction, and the tiffany-blue space represents correct prediction. }
\label{fig.examples}
\end{center}
\end{figure}

\noindent\textbf{Frame-level Metric.} Pixel-level metric is good to evaluate the uncertainty estimation. However, pixel-wise uncertainty estimation is hard to leverage in real applications. For example, active learning system wants to find which frame is valuable to be labeled rather than decides which pixel should be labeled. Here, we propose frame-level uncertainty metrics to show that our uncertainty estimation can work well and faster in real applications. The procedure is shown in Fig.~\ref{fig.frame_level_explain}. First, frames are ranked by the error of prediction as the ground truth ranking sequence. Then, we rank frames by the uncertainty estimation and evaluate the uncertainty ranking sequence by two metrics: \textbf{Kendall tau} \cite{kendall1938new} and \textbf{Ranking IoU}. Kendall tau is a well-known ranking metric measures how the ranking sequence is similar to the ground truth sequence. The value is bounded in 1 (fully identical sequence) and -1 (fully different sequence). Table~\ref{table.frame_rank} shows the comparison of Kendall tau by SegNet and Tiramisu backbone.Though Table~\ref{table.frame_rank} shows that the highest scores appears in \textit{Mean STD} and \textit{BALD}, RTA-MC outperforms MC in \textit{Entropy} and \textit{Variation Ratio}. It also shows that RTA-MC improves frame-level ranking compare to TA-MC. It is attributed to the decision function $A(E)$ which reduces the uncertainty value of pixels with wrong flow estimation which harm the frame-level uncertainty. Kendall tau compares the whole sequence similarity, but real applications pay more attention on higher ranking similarity than whole ranking. Therefore, we define a novel frame-level ranking metric called Ranking IoU. Given a percentage of frame $P_{f}$ to retrieve, we retrieve frames depend on error $G\left(P_{f}\right) = \left \{ g_{1}, g_{2}, ..., g_m, ..., g_{M} \right \}$ and uncertainty $U\left(P_{f}\right) = \left \{ u_{1}, u_{2}, ..., u_m, ..., u_{M} \right \}$. The ranking IoU is:
\begin{align}
\text{Ranking IoU} = \frac{G\left ( P_{f}  \right )\cap U\left ( P_{f}  \right )}{G\left ( P_{f}  \right )\cup  U\left ( P_{f}  \right )}
\end{align} 
Larger Ranking IoU means that those frames we choose are hard to predict, so they are valuable to be labeled. Left of the Table~\ref{table.rankingIOU} shows the ranking IoU performance of SegNet backbone between different methods and uncertainty functions. We show performance in different $P_{f}$. In column $P_{f}=10\%$, TA in \textit{Variation Ratio} has $52.2\%$ which is larger than RTA's $47.8\%$ about $4.4\%$; however, $10\%$ of test data only contains 23 frames so that RTA and TA actually only have 1 frame difference. For $P_{f}=30\%$ which is a practical percentage for real applications, RTA outperforms other methods in all uncertainty functions. The best score of RTA $69.6\%$ is larger than MC dropout's best score $66.7\%$ about $3\%$, which means our uncertainty method can more retrieve $3\%$ of hardest frames. For \textit{Entropy} and \textit{Variation Ratio}, RTA outperforms other methods in almost all percentage. The right of the Table~\ref{table.rankingIOU} shows the Ranking IOU of tiramisu backbone. The results are similar to SegNet backbone that for \textit{Entropy} and \textit{Variation Ratio}, RTA outperforms other methods. Table~\ref{table.frame_rank} and Table~\ref{table.rankingIOU} indicate that RTA can generate high-quality uncertainty. Fig.~\ref{fig.examples} shows the visualization of prediction, uncertainty, and error. It shows that RTA's uncertainty quality is comparable to MC dropout and the large uncertainty pixels are correlated to the misclassified pixels. 


\begin{table}[t]
\centering
\addtolength{\tabcolsep}{2mm}
\begin{tabular}{lcccc}
\hline
Method & Entropy & Variation Ratio & Mean STD & BALD  \\ 
\hline\hline
SegNet MC   &  0.647  &  0.668  &\textbf{0.677} & \textbf{0.671}\\ 
SegNet TA-MC   &  0.626  &  0.631 & 0.540 & 0.527\\ 
SegNet RTA-MC & \textbf{0.662} & \textbf{0.674} & 0.627 & 0.620\\
\hline\hline
Tiramisu MC   &  0.636  &  0.653  & 0.659 & \textbf{0.647}\\ 
Tiramisu TA-MC   &  0.660 &  0.674 & \textbf{0.663} & 0.626\\ 
Tiramisu RTA-MC & \textbf{0.664} & \textbf{0.678} & 0.635 & 0.612\\
\hline
\end{tabular}
\small \caption{\small \New{Comparison of Kendall tau. For both backbone, the result shows that RTA-MC has the highest value in \textit{Entropy} and \textit{Variation Ratio}. For overall performance, RTA-MC is comparable to MC dropout. }}
\label{table.frame_rank}
\end{table}

\label{sec.unct}

\begin{table}[t]
\centering
\scriptsize
\setlength\belowcaptionskip{-20pt}
\begin{minipage}{.5\textwidth}
\begin{tabular}{ccllll}
\hline
\multirow{2}{*}{Metric}          & \multirow{2}{*}{Method} & \multicolumn{4}{c}{Percentage}                                                                            \\
                                 &                         & \multicolumn{1}{c}{10\%} & \multicolumn{1}{c}{30\%} & \multicolumn{1}{c}{50\%} & \multicolumn{1}{c}{70\%} \\ \hline\hline
\multirow{3}{*}{Entropy}         & MC                      & 47.8                     & 59.4                     & 72.4                     & 85.2                     \\
                                 & TA                      & 47.8                     & 65.2                     & 69.8                     & 85.8                     \\
                                 & RTA                     & 47.8                     & \textbf{66.7}            & \textbf{73.3}            & \textbf{88.9}            \\ \hline
\multirow{3}{*}{Variation Ratio} & MC                      & 47.8                     & 62.3                     & 74.1                     & 85.2                     \\
                                 & TA                      & \textbf{52.2}            & 63.8                     & 70.7                     & 85.2                     \\
                                 & RTA                     & 47.8                     & \textbf{65.2}            & \textbf{76.7}            & \textbf{88.3}            \\ \hline
\multirow{3}{*}{Mean STD}        & MC                      & \textbf{52.2}            & 66.7                     & \textbf{71.6}            & \textbf{86.4}            \\
                                 & TA                      & 43.5                     & 65.2                     & 65.5                     & 74.1                     \\
                                 & RTA                     & 43.5                     & \textbf{69.6}            & 69.8                     & 82.7                     \\ \hline
\multirow{3}{*}{BALD}            & MC                      & \textbf{47.8}            & 66.7                     & \textbf{71.6}            & \textbf{87.7}            \\
                                 & TA                      & 43.5                     & 62.3                     & 64.7                     & 74.1                     \\
                                 & RTA                     & 43.5                     & \textbf{69.6}            & 67.2                     & 82.1                      \\ \hline
\end{tabular}
  
  \centering
  \label{table.rankingIOU_segnet}
\end{minipage}%
\begin{minipage}{.5\textwidth}
 \begin{tabular}{ccllll}
\hline
\multirow{2}{*}{Metric}          & \multirow{2}{*}{Method} & \multicolumn{4}{c}{Percentage}                                                                            \\
                                 &                         & \multicolumn{1}{c}{10\%} & \multicolumn{1}{c}{30\%} & \multicolumn{1}{c}{50\%} & \multicolumn{1}{c}{70\%} \\ \hline\hline
\multirow{3}{*}{Entropy}         & MC                      & 34.8                     & 60.9                     & 70.7                     & \textbf{86.4}                     \\
                                 & TA                      & \textbf{47.8}                     & \textbf{63.8}                     & 71.6                     & 84.6                     \\
                                 & RTA                     & \textbf{47.8}                     & \textbf{63.8}            & \textbf{74.1}            & \textbf{86.4}            \\ \hline
\multirow{3}{*}{Variation Ratio} & MC                      & 34.8                     & 60.9                     & 74.1                     & 86.4                     \\
                                 & TA                      & 47.8            & \textbf{65.2}                     & 72.4                     & 85.8                     \\
                                 & RTA                     & \textbf{52.1}                     & \textbf{65.2}            & \textbf{75.9}            & \textbf{87.7}            \\ \hline
\multirow{3}{*}{Mean STD}        & MC                      & 30.4            & 63.8                     & \textbf{76.7}            & \textbf{86.4}            \\
                                 & TA                      & \textbf{47.8}                     & \textbf{75.4}                     & 74.1                     & 82.0                     \\
                                 & RTA                     & 43.4                     & 68.1            & 73.3                     & 82.1                     \\ \hline
\multirow{3}{*}{BALD}            & MC                      & 30.4            & 62.3                     & 72.4            & \textbf{86.4}            \\
                                 & TA                      & 43.5                     & \textbf{71.0}                     & \textbf{71.6}
                                & 80.2                     \\
                                 & RTA                     & \textbf{47.8}                     & 66.7            & \textbf{71.6}                     & 80.9                      \\ \hline
\end{tabular}
  
  \centering
  \label{table.rankingIOU_tiramisu}
\end{minipage}
\small \caption{\small Ranking IoU. Left table is the result of SegNet backbone. Right table is the result of Tiramisu backbone. For retrieving 30\%, 50\% and 70\% of frames RTA-MC have the highest score by using Variation ratio.}
\label{table.rankingIOU}
\end{table}

\section{Conclusions}
In this work, we propose the region-based temporal aggregation (RTA) method to simulate the sampling procedure of Monte Carlo (MC) dropout for video segmentation. Our RTA method utilizes the temporal information from videos and only needs to sample one time to generate the prediction and the uncertainty for each frame. Compared to using general MC dropout, RTA can achieve similar performance on CamVid dataset with only 1.2\% drop on mean IoU metric and incredibly speed up the inference process 10.97 times. Moreover, the uncertainty obtained by the RTA method is also comparable on pixel-level metric and even outperforms MC dropout on frame-level metric when using \textit{Entropy} and \textit{Variation Ratio} as the uncertainty estimation function. With our faster approach, we expect to extend our method on instance segmentation task in future work. In real-time applications, it's more important to obtain the instance-level uncertainty more precisely. 
\section{Acknowledgment}
We thank Umbo CV, MediaTek, MOST 107-2634-F-007-007 for their support.

\clearpage

\bibliographystyle{splncs04}
\bibliography{egbib}
\end{document}